\documentclass{article} % For LaTeX2e
\usepackage{template/iclr2026_conference,times}
\usepackage{template/packages}

\newcommand{\glass}{\raisebox{-0.2\height}{\protect  \includegraphics[height=1em]{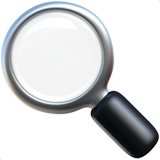}}}
\title{\centering \glass \hspace{0.1em} Layer by layer, module by module: \\ Choose both for optimal OOD probing of ViT
}

\def\authspace{1.5}
\def\authbspace{1}
\author{\\
  \makebox[\textwidth][c]{ % This centers everything relative to the page margins
    \begin{tabular}{cc}
      \textbf{Ambroise Odonnat}\thanks{Correspondence to \href{mailto:ambroise.odonnat@gmail.com}{\texttt{ambroise.odonnat@gmail.com}}} \hspace{\authspace cm} & \textbf{Vasilii Feofanov} \\
      Noah's Ark Lab, Inria \hspace{\authspace cm} & 42.com 
      \\[1ex] 
    \end{tabular}
  }
  \\ \\
  \makebox[\textwidth][c]{
    \begin{tabular}{ccc}
      \textbf{Laetitia Chapel} \hspace{\authbspace cm} & \textbf{Romain Tavenard} \hspace{\authbspace cm} & \textbf{Ievgen Redko} \\
      L'Institut Agro, IRISA \hspace{\authbspace cm} & Univ. Rennes 2, IRISA \hspace{\authbspace cm} & Noah's Ark Lab
    \end{tabular}
  }
}

\iclrfinalcopy
\begin{document}

\maketitle

\begin{abstract}
Recent studies have observed that intermediate layers of foundation models often yield more discriminative representations than the final layer. While initially attributed to autoregressive pretraining, this phenomenon has also been identified in models trained via supervised and discriminative self-supervised objectives. In this paper, we conduct a comprehensive study to analyze the behavior of intermediate layers in pretrained vision transformers. Through extensive linear probing experiments across a diverse set of image classification benchmarks, we find that distribution shift between pretraining and downstream data is the primary cause of performance degradation in deeper layers. Furthermore, we perform a fine-grained analysis at the module level. Our findings reveal that standard probing of transformer block outputs is suboptimal; instead, probing the activation within the feedforward network yields the best performance under significant distribution shift, whereas the normalized output of the multi-head self-attention module is optimal when the shift is weak. 
\begin{center}
\faGithub \quad \href{https://github.com/ambroiseodt/vit-probing}{\texttt{vit-probing}}
\end{center}

\end{abstract}

\section{Introduction}
Foundation models, which rely primarily on the transformer architecture~\citep{vaswani2017attention}, have achieved impressive performance in a wide range of areas such as natural language processing~\citep{touvron2023llama,brown2020gpt3}, computer vision~\citep{simeoni2025dinov3}, time series forecasting~\citep{nie2023patchtst, ilbert2024samformer}, and mathematical reasoning~\citep{comanici2025gemini25pushingfrontier}. These models are typically pretrained on massive amount of diverse data~\citep{kimiteam2025kimivltechnicalreport, shukor2025scalinglawsoptimaldata} to encode general knowledge and then adapted to downstream tasks via finetuning~\citep{lee2023surgical,hu2022lora} or used as frozen feature extractors~\citep{oquab2024dinov,el_nouby2025aim} (note that the goal is different in tool-augmented large language models where models learn to use a tool instead of incorporating knowledge in their weights~\citep{lewis2020rag, houliston2025provable, schick2023toolformer}). However, the reliability of foundation models remains a critical challenge in real environments. As they operate at scale, distribution shifts inevitably occur, and hidden representations remain effective only if the knowledge integrated during pretraining provides a robust prior that withstands drift at deployment.

\paragraph{Representations under drift.} In practice, distribution shifts can severely degrade the performance by making the features extracted on out-of-distribution (OOD) data uninformative~\citep{quionero2009shift}. For foundation models, where pretraining data is often inaccessible, detecting and responding to this drift is a particularly challenging task. Notably, while some approaches assume the type of shift is known a priori~\citep{lee2023surgical, xie2024mano, xie2025leveraging,uselis2025intermediate}, this assumption rarely holds for general-purpose models. In this setting,~\citet{skean2025layer} challenged the conventional wisdom that final-layer representations are universally optimal, demonstrating that intermediate layers can yield superior performance. In particular, they argued that autoregressive vision models benefit from intermediate layers, whereas the final layer remains optimal for vision transformers~\citep[ViT,][]{dosovitskiy2021vit}. Other recent works appear to contradict this conclusion~\citep{uselis2025intermediate,bolya2025perception}, necessitating further investigation.

\paragraph{Our approach.} In this paper, we study pretrained ViTs on out-of-distribution downstream image classification tasks and investigate when and why intermediate layers outperform the final layer in a linear probing setup (\cref{sec:layer_by_layer}). We identify the distribution shift between pretraining and downstream data as the driving factor of this phenomenon, finding intermediate layers to be significantly more robust than the final ones. Motivated by prior works on component-wise adaptation~\citep{zhao2024tuning,odonnat2026vitplasticity}, we conduct a fine-grained study by probing each type of transformer module: normalization layers, multi-head attention modules, residual connections, and feedforward layers\footnote{In what follows, we use the terms module and component interchangeably, always referring to normalization layers, multi-head attention modules, residual connections, and feedforward layers.} (\cref{sec:module_by_module}). We find that transformer modules are not equally resilient to the shift. 
\begin{tcolorbox}[colback=\boxcol,
    colframe=black,
    arc=4pt,
    boxsep=0.3pt,
    boxrule=\boxwidth pt,
]%
\textbf{Takeaways.} Our analysis reveals two \textit{actionable} takeaways summarized below:
\begin{enumerate}[leftmargin=*]
    \item In ID settings, final layers always yield better performance than intermediate layers;
    \item  In OOD settings, probing inputs and activations of intermediate feedforwards is better.
\end{enumerate}
\end{tcolorbox}

\section{Experimental setup}
\label{sec:exp_setup}
\paragraph{Vision transformer.} A ViT takes as input 2D images, which are split into square patches and fed to a succession of transformer encoders. As shown in \cref{fig:transformer_encoder}, each block consists of alternating multi-head attention modules  (\textbf{MHA}) and feedforward networks. The latter combines two fully connected layers, \textbf{FC1} ($d\to 4d$) and \textbf{FC2} ($4d\to d$), separated by a GeLU activation \citep[\textbf{Act},][]{hendrycks2016gelu}. Two LayerNorms~\citep{ba2016layernormalization}, \textbf{LN1} and \textbf{LN2}, precede the MHA and FFN, and two residual connections, \textbf{RC1} and \textbf{RC2}, follow them. In this work, we track the outputs of the $8$ operations within each layer, denoting them by the name of their corresponding module. Note that the standard approach is to probe \textbf{RC2}, the output of the transformer block~\citep{oquab2024dinov,simeoni2025dinov3,el_nouby2025aim}.

\paragraph{Implementation.} All our experiments are conducted with an $86$M-parameter ViT pretrained on ImageNet-21k~\citep{deng2009imagenet}. For a given hidden representation, linear probing is done by pooling the embeddings of the \texttt{CLS} token and applying a logistic regression with the \texttt{L-BFGS} solver. We consider a diverse set of $11$ commonly used classification benchmarks: Cifar10, Cifar100~\citep{krizhevsky2009learning}; $5$ variants from Cifar10-C~\citep{hendrycks2019robustness}: Contrast, Gaussian Noise, Motion Blur, Snow, Speckle Noise; $2$ domains from DomainNet~\citep{peng2019moment}: Clipart, Sketch; Flowers102~\citep{nilsback2008flowers102} and Pets~\citep{parkhi2012pets}. The preprocessing protocol follows~\citet{dosovitskiy2021vit}. The full implementation details are given in~\cref{app:implem}.

\section{Distribution shift degrades the performance of final layers}
\label{sec:layer_by_layer}
\citet{skean2025layer} showed that intermediate layers of large language models consistently outperform the final ones. When conducting a similar analysis on vision transformers, they observe increasing downstream performance toward the final layers, except for the \textsc{Aim} model~\citep{el_nouby2025aim}, which is the only autoregressive vision model. It leads them to conclude that the benefit of intermediate layers is not modality-dependent but rather a byproduct of pretraining, with the autoregression as the driving factor. 

\paragraph{Motivation.} We notice that the vision experiments performed by \citet{skean2025layer} are limited to ImageNet~\citep{deng2009imagenet}, which is included in the pretraining set of the models studied~\citep{dosovitskiy2021vit,oquab2024dinov,bao2022beit,he2022mae}. We go beyond this in-distribution (ID) setting and perform linear probing on a diverse set of out-of-distribution (OOD) downstream data. As a sanity check, we conduct the same experiment in the ID scenario by finetuning the pretrained model on each dataset, respectively. The training protocol follows~\citet{dosovitskiy2021vit}, see~\cref{app:implem} for details.

\begin{figure}[!h]
    \centering
    \includegraphics[width=\linewidth]{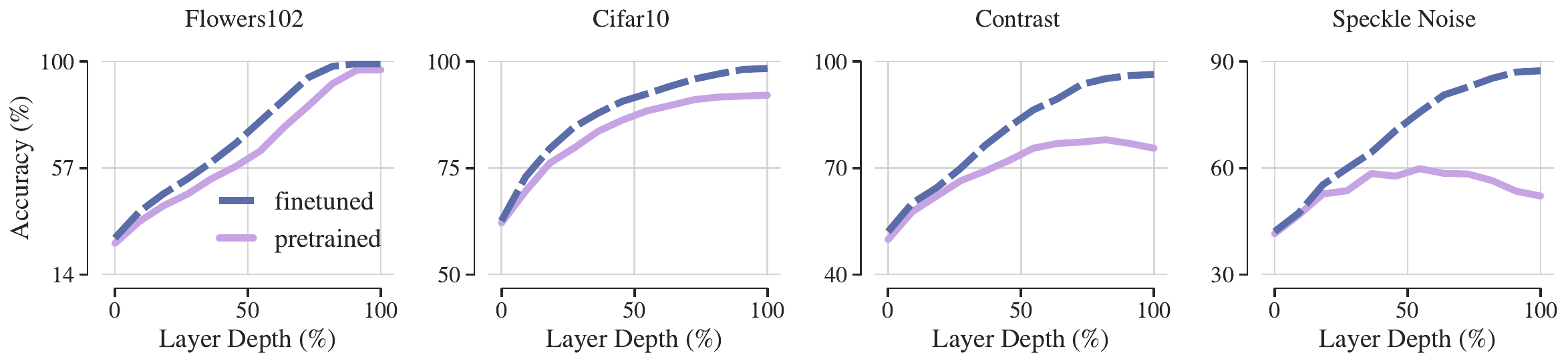}
    \caption{\textbf{Layer by layer.} Evolution of the linear probing performance across each layer of an $86$M ViT pretrained on ImageNet (the x-axis is the depth percentage of the layer). The solid line denotes the model only pretrained, and the dashed line denotes the model finetuned on the dataset at hand. From \textbf{left} to \textbf{right}, the shift between the pretraining and the downstream data increases. The stronger the shift, the worse the final layers.}
    \label{fig:layer_by_layer}
\vspace{-5mm}
\end{figure}

\paragraph{Results.}  In~\cref{fig:layer_by_layer}, the linear probing performance across layers is displayed for the pretrained model (solid line) and for the finetuned model (dashed line). For each layer, probing is done on \textbf{RC2}, the output of the transformer block. As we have no information on the degree of distribution shift between pretraining and downstream data, we hypothesize that the stronger the shift, the larger the performance gap between frozen and finetuned encoders. We sort the plots in decreasing order of finetuning performance (see~\cref{tab:finetuning_results}) from left to right: Flowers102, Cifar10, Contrast, and Speckle Noise. Our findings are twofold: \textbf{(1)} in the OOD scenario, we observe that the deeper representations become worse as the shift increases from left to right; \textbf{(2)} conversely, in the ID setting, the best visual embedding is at the end of the network. Similar patterns can be observed on all datasets in~\cref{fig:layer_by_layer_all}. 
\paragraph{Distribution shift as the culprit.} From \cref{fig:layer_by_layer}, we conclude that the intermediate layers are more robust to distribution shifts than final layers. This can be intuitively understood by the fact that layers tend to specialize closer to the classification head. Our findings show that the benefit of intermediate representations is not merely a byproduct of the pretraining objective, as suggested by~\citet{skean2025layer}, but also a consequence of the eventual presence of distribution shift. As such, when finetuning is prohibitive, being able to identify whether the setting is ID or OOD is crucial to know which accuracy profile to expect (from left to right in~\cref{fig:layer_by_layer}) and which layer to probe. 

\section{Not all transformer modules are worth probing on OOD data}
\label{sec:module_by_module}
\begin{figure}[!b]
\vspace{-5mm}
    \centering
    \begin{minipage}[c]{0.3\textwidth}
        \centering
        \begin{adjustbox}{width=.8\linewidth}
        \begin{tikzpicture}[scale=1,baseline={([yshift=-.5ex]current bounding box.center)}]
        \begin{scope}[xshift=-3cm]
\definecolor{ln1_color}{HTML}{daa4ac} 
\definecolor{mha_color}{HTML}{37abb5}
\definecolor{rc1_color}{HTML}{d0e2c2}
\definecolor{ln2_color}{HTML}{b153a1}
\definecolor{ffn_color}{RGB}{194,232,247}
\definecolor{fc1_color}{HTML}{a291e1}
\definecolor{act_color}{HTML}{96c0cf}
\definecolor{fc2_color}{HTML}{858ec2}
\definecolor{rc2_color}{HTML}{428379}
\definecolor{gray_bbox_color}{RGB}{247,247,247}
\def\background{gray_bbox_color}

% The bounding box for the Encoder Block (Unchanged)
\draw[fill=\background, line width=0.046875cm, rounded corners=0.300000cm] (-0.55000, -0.45) -- (2.8000, -0.45) -- (2.8000, 6.12) -- (-0.55000, 6.12) -- cycle;

% Input (Unchanged)
\draw[line width=0.046875cm, -latex] (1.250000, -0.950000) -- (1.250000, 0.300000);

% LN1 (Unchanged)
\draw[line width=0.046875cm, fill=ln1_color!80, rounded corners=0.100000cm] (0.450000, 0.800000) -- (2.0000, 0.800000) -- (2.0000, 0.300000) -- (0.450000, 0.300000) -- cycle;
\node[text width=2.500000cm, align=center] at (1.250000,0.550000) {\textbf{LN1}};
\draw[line width=0.046875cm, -latex] (1.250000, 0.800000) -- (1.250000, 1.60000);

% Skip Connections for MHA (Unchanged)
\draw[-latex, line width=0.046875cm, rounded corners=0.200000cm] (1.250000, -0.20000) -- (-0.25000, -0.2000) -- (-0.250000, 2.720000) -- (1.05000, 2.720000);
\draw[-latex, line width=0.046875cm, rounded corners=0.200000cm] (1.250000, 1.230000) -- (0.7000, 1.230000) -- (0.7000, 1.63000);
\draw[-latex, line width=0.046875cm, rounded corners=0.200000cm] (1.250000, 1.230000) -- (1.80, 1.230000) -- (1.80, 1.63000);

% MHA (Unchanged)
\draw[line width=0.046875cm, fill=mha_color!90, rounded corners=0.100000cm] (0.450000, 2.130000) -- (2.0000, 2.130000) -- (2.0000, 1.630000) -- (0.450000, 1.630000) -- cycle;
\node[text width=2.500000cm, align=center] at (1.250000,1.880000) {\textbf{MHA}};
\draw[line width=0.046875cm] (1.250000, 2.130000) -- (1.250000, 2.5250000);
\draw[line width=0.046875cm, -latex] (1.250000, 2.940000) -- (1.250000, 3.430000);

% Add (Unchanged)
\node[circle, draw, minimum size=0.25em, fill=rc1_color!80, inner sep=0pt, line width=0.046875cm, align=center, label=right:{\textbf{RC1}}] at (1.250000,2.720000) {$\bm{\mathbf{+}}$};

% LN2 (Unchanged)
\draw[line width=0.046875cm, fill=ln2_color!70, rounded corners=0.100000cm] (0.450000, 3.930000) -- (2.00000, 3.930000) -- (2.00000, 3.430000) -- (0.450000, 3.430000) -- cycle;
\node[text width=2.500000cm, align=center] at (1.250000,3.68000) {\textbf{LN2}};
\draw[line width=0.046875cm, -latex] (1.250000, 3.93000) -- (1.250000, 4.68000);

% Skip Connections for FFN
\draw[-latex, line width=0.046875cm, rounded corners=0.200000cm] (1.250000, 3.050000) -- (-0.250000, 3.050000) -- (-0.250000,5.770000) -- (1.05000, 5.770000);

% FFN
\draw[line width=0.046875cm, fill=white, rounded corners=0.100000cm] (0.450000, 5.18000) -- (2.00000, 5.18000) -- (2.00000, 4.68000) -- (0.450000, 4.68000) -- cycle;
\node[text width=2.500000cm, align=center] at (1.250000,4.930000) {FFN};
\draw[line width=0.046875cm] (1.250000, 5.180000) -- (1.250000, 5.5750000);
\draw[line width=0.046875cm, -latex] (1.250000, 5.990000) -- (1.250000, 6.480000);

% Add (Unchanged)
\node[circle, draw, minimum size=0.25em, fill=rc2_color!60, inner sep=0pt, line width=0.046875cm, align=center, label=right:{\textbf{RC2}}] at (1.250000,5.770000) {$\bm{\mathbf{+}}$};

% Zoom
\draw[thick][dashed](2.0, 4.68) -- (3.05, 3.87);
\draw[thick][dashed] (2.0, 5.18) -- (3.05, 6.1);

% FC1 (Unchanged)
\draw[line width=0.046875cm, fill=fc1_color!80, rounded corners=0.100000cm] (3.050000, 4.34) -- (4.6000, 4.34) -- (4.6000, 3.84) -- (3.050000, 3.84) -- cycle;
\node[text width=2.500000cm, align=center] at (3.850000,4.090000) {\textbf{FC1}};
\draw[line width=0.046875cm] (3.850000, 4.34) -- (3.850000, 4.685000);
\draw[line width=0.046875cm, -latex] (3.850000, 5.19000) -- (3.850000, 5.64000);

% Sigma Block 
\node[circle, draw, minimum size=1.2em, fill=act_color!60, inner sep=0pt, line width=0.046875cm, align=center, label=right:{\textbf{Act}}] at (3.850000,4.930000) {$\bm{\sigma}$};

% FC2 
\draw[line width=0.046875cm, fill=fc2_color!70, rounded corners=0.100000cm] (3.050000, 6.1400) -- (4.60000, 6.14000) -- (4.60000, 5.64000) -- (3.050000, 5.64000) -- cycle;
\node[text width=2.500000cm, align=center] at (3.850000,5.890000) {\textbf{FC2}};
    % All your existing diagram code here
\end{scope}
        \end{tikzpicture}
        \end{adjustbox}
        \captionof{figure}{\bfseries Transformer block.}
        \label{fig:transformer_encoder}
    \end{minipage}
    \hfill
    \begin{minipage}[c]{0.69\textwidth}
        \centering
        \captionof{table}{\textbf{Module by module.} For each module, we report the best linear probing accuracy over the layers. The best performance per dataset is in \textbf{bold} and the module with the highest win rate is in \colorbox{tablegray}{gray}.
            }
        \scalebox{.7}{
        \begin{NiceTabular}{l|cccccccc}
        % Color rows
        \CodeBefore
            \columncolor{tablegray}{7} 
            \rowcolor{white}{1}
        \Body
         Dataset & \textbf{LN1} & \textbf{MHA} & \textbf{RC1} & \textbf{LN2} & \textbf{FC1} & \textbf{Act} & \textbf{FC2} & \textbf{RC2} \\
         \toprule[\thick pt] 
        Cifar10 & 91.94 & 91.98 & 92.19 & 92.20 & \textbf{92.28} & 85.30 & 89.98 & 92.07\\
        Cifar100 & 69.39 & 67.27 & 69.88 & \textbf{69.97} & 68.98 & 69.75 & 60.92 & 69.63\\
        Contrast & 78.05 & 74.55 & 78.30 & 79.05 & 78.15 & \textbf{80.20} & 70.85 & 77.95\\
        Gaussian Noise & 57.65 & 60.40 & 59.05 & 58.10 & 58.75 & \textbf{61.85} & 56.70 & 58.20\\
        Motion Blur & 66.85 & 64.30 & 68.50 & 67.75 & 66.80 & \textbf{71.15} & 60.90 & 67.75\\
        Snow & 67.30 & 66.15 & 68.10 & 67.30 & 67.35 & \textbf{69.30} & 61.50 & 67.70\\
        Speckle Noise & 59.75 & 61.85 & 60.25 & 59.95 & 59.90 & \textbf{63.35} & 58.00 & 59.80\\
        Clipart & 47.66 & 43.82 & 48.66 & 48.37 & 45.74 & \textbf{49.34} & 40.97 & 48.33\\
        Sketch & 32.36 & 30.95 & 33.32 & 33.07 & 31.11 & \textbf{34.90} & 28.45 & 32.99\\
        Flowers102 & 96.58 & 96.34 & 96.58 & \textbf{96.62} & 96.44 & 91.64 & 95.23 & \textbf{96.62}\\
        Pet & 88.36 & 88.33 & 89.48 & \textbf{89.51} & 88.47 & 83.46 & 85.80 & 89.18\\
        \end{NiceTabular}
        }
        \label{tab:linear_probing_results}
    \end{minipage}
\end{figure}
In the previous section, we observed that the linear probing performance of the last transformer layers tends to degrade under distribution shifts. Inspired by~\citet{odonnat2026vitplasticity}, where the authors demonstrated that the transformer modules do not adapt to downstream data equally, we investigate whether probing after a specific module within a transformer layer has an impact on the performance. We conduct a similar but finer-grained analysis to that in~\cref{sec:layer_by_layer} on a broader set of downstream data across \textbf{both layers and modules} of the $86$M ViT pretrained on ImageNet~\citep{deng2009imagenet}. 

\begin{figure}[!h]
    \centering
    \includegraphics[width=\linewidth]{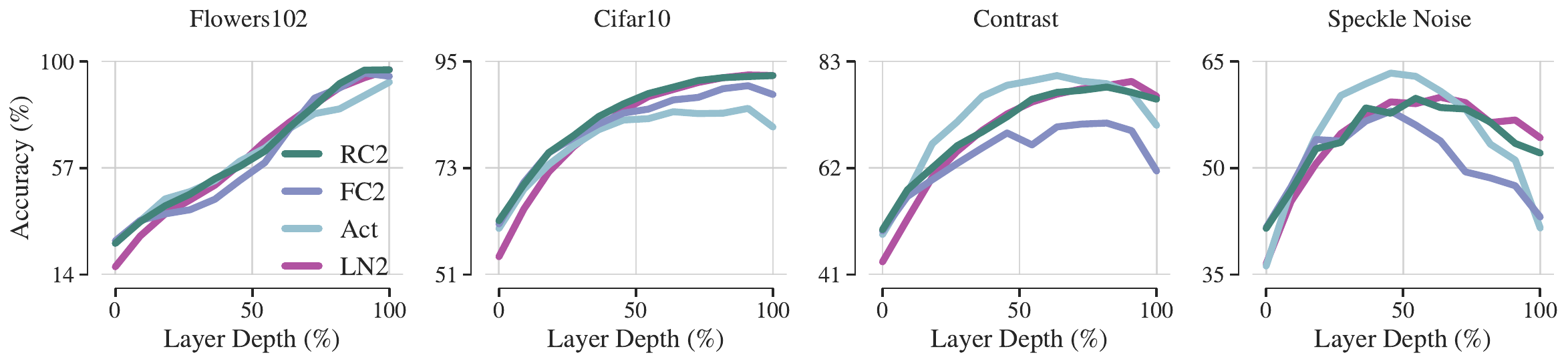}
    \caption{\textbf{Layer by layer, module by module.} Evolution of the linear probing performance of transformer modules across the layers of an $86$M ViT pretrained on ImageNet. From \textbf{left} to \textbf{right}, the shift between the pretraining and the downstream data increases.}
    \label{fig:module_by_module}
\end{figure}

\paragraph{Results.} In~\cref{tab:linear_probing_results}, we report for each dataset-module pair the best linear probing accuracy achieved across the layers. Our findings are threefold: \textbf{(1)} the standard probing on transformer block outputs (\textbf{RC2}) is suboptimal on all datasets but one, Flowers102, for which most components perform equally well; \textbf{(2)} \textbf{FC2} is the worst module to probe, with the lowest accuracy on 10 out of 12 datasets, while \textbf{Act} is the best-performing one overall, with the highest win-rate over all datasets. We note that it outperforms other components by a large margin when the shift is strong, despite being less good on easier datasets such as Cifar10, Flowers102, or Pet. \textbf{(3)} Other modules yield comparable results, with \textbf{LN2} being slightly above. We further study these modules along the depth in~\cref{fig:module_by_module}. Akin to~\cref{fig:layer_by_layer}, the plots are ordered in terms of increasing distribution shift between pretraining and downstream data: Flowers102, Cifar10, Contrast, and Speckle Noise. We can see that \textbf{FC2} consistently yields the worst performance. When the shift increases, \textbf{Act} is the best module in intermediate layers while its performance in final layers plunges. On the contrary, \textbf{LN2} and \textbf{RC2} yield subpar but more stable results. Similar patterns can be observed on all datasets in~\cref{fig:module_by_module_all}. 
\paragraph{Signal propagation.} We found in~\cref{sec:layer_by_layer}, intermediate representations are more robust to distribution shifts than final layers. As the residual stream of each encoder bears information from the previous layers, this could explain why the accuracy profiles of \textbf{LN2} and \textbf{RC2} are less concave than \textbf{Act} and \textbf{FC2}. Furthermore, all modules are maps from $(\RR^d)^n$ to $(\RR^d)^n$, except the feedforward network, where \textbf{FC1} increases the dimension of tokens to $4d$ and \textbf{FC2} decreases back to $d$. We hypothesize that by operating in a higher dimension, \textbf{FC1} and \textbf{Act} help promote feature disentanglement, which would benefit the probing. Since \textbf{Act} filters the potential noise induced by the projection, it may explain its higher accuracy. Conversely, \textbf{FC2} compresses the input, which may impact the linear separability of data. Another interesting perspective comes from seeing feedforward networks as key-value memory~\citep{geva2021ffn}. The observed behavior can then be understood by the fact that \textbf{FC1} and \textbf{Act} capture semantic information in the inputs, which can be useful for linear probing, while \textbf{FC2} merely reflects a distribution over tokens. Our findings motivate further study of the hidden representations of transformers. A promising direction would be to extend the analysis of~\citet{skean2025layer} with information-theoretic, geometric, and invariance measures at the level of transformer modules. A key takeaway for practitioners is that probing after the activation might lead to the best performance, provided the correct layer is chosen. A safer approach, if the shift is difficult to detect, is to probe the \textbf{LN2} module, rather than the standard choice of \textbf{RC2}. 

\section{Discussion}
We study linear probing on pretrained vision transformers across both layers and modules on a diverse set of classification benchmarks. We find that the discrepancy between pretraining and downstream data is at fault for the degradation of the final layers' performance, while intermediate representations are more robust. We further notice that the standard choice of probing the outputs of transformer blocks is not optimal. In comparison, the hidden representations after the feedforward activation in intermediate layers are the richest under significant distribution shift, whereas the outputs of the LayerNorm preceding the feedforward network are better when the shift is almost negligible. Our work provides a novel perspective on the role of vision transformer hidden representations. We hope it will help guide efficient methods towards detecting distribution shifts and identifying the layers and modules to probe.

\newpage
\bibliography{references}
\bibliographystyle{template/iclr2026_conference}

\newpage
\appendix
\textbf{\LARGE Appendix}
\section{Implementation details}
\label[secinapp]{app:implem}
\paragraph{Vision transformers.} In vision transformers~\citep[ViT,][]{dosovitskiy2021vit}, input $2$D images are split into square patches of size $P$, which are then flattened and linearly embedded into dimension $d$. A classification token (\texttt{CLS}) is prepended to the sequence of patch tokens before positional embeddings are added. The obtained sequence $\mathbf{x} = (\mathbf{x}_1, \ldots, \mathbf{x}_n) \in \RR^{d\times n}$ is fed through a succession of transformer layers~\citep{vaswani2017attention}, where the output representation of the \texttt{CLS} token serves as the final encoder output. In our code, we follow the original ViT implementation from~\citet{dosovitskiy2021vit} and use a convolutional layer to embed images~\citep[see][\S ``Hybrid Architecture"]{dosovitskiy2021vit}. This is also the standard in the implementation from~\citet{transformers}. In~\cref{fig:vit_implementation}, we display the implementation of the ViT-Base model with a classification head for $10$ classes (we renamed our package ``my\_lib" to respect the anonymity).

\begin{figure}[!h]
\begin{center}
\begin{minipage}{0.95\linewidth}
\begin{lstlisting}[language=Python, frame=single]
# Python snippet to print the ViT architecture
from my_lib import ViT

model = ViT(name="base", n_classes=10)
print(model)

# Corresponding output
Transformer(
  (embedding): Embedding(
    (patching): PatchImages(
      (patching): Sequential(
        (0): Conv2d(3, 768, kernel_size=(16, 16), stride=(16, 16))
        (1): Flatten(start_dim=2, end_dim=-1)
      )
    )
  )
  (blocks): ModuleList(
    (0-11): 12 x TransformerBlock(
      (attn_norm): LayerNorm((768,), eps=1e-12, elementwise_affine=True)
      (attn): SelfAttention(
        (qkv_mat): Linear(in_features=768, out_features=2304, bias=True)
        (output): Linear(in_features=768, out_features=768, bias=True)
      )
      (ffn_norm): LayerNorm((768,), eps=1e-12, elementwise_affine=True)
      (ffn): FeedForward(
        (fc1): Linear(in_features=768, out_features=3072, bias=True)
        (fc2): Linear(in_features=3072, out_features=768, bias=True)
      )
    )
  )
  (output): Output(
    (output_layer): ClassificationLayer(
      (output_norm): LayerNorm((768,), eps=1e-12, elementwise_affine=True)
      (output): Linear(in_features=768, out_features=10, bias=True)
    )
  )
)
\end{lstlisting}
\end{minipage}
\end{center}
\caption{ViT-Base Implementation.}
\label{fig:vit_implementation}
\end{figure}

\paragraph{Data preprocessing.} All our experiments are conducted on a varied collection of $11$ classification benchmarks: Cifar10, Cifar100~\citep{krizhevsky2009learning}; variants from Cifar10-C~\citep{hendrycks2019robustness} with severity $5$: Contrast, Gaussian Noise, Motion Blur, Snow, Speckle Noise; $2$ domains from DomainNet~\citep{peng2019moment}, a challenging benchmark typically used for domain generalization: Clipart, Sketch; Flowers102~\citep{nilsback2008flowers102} and Pets~\citep{parkhi2012pets}. The preprocessing follows~\citet{dosovitskiy2021vit} and~\citet{kolesnikov2020bit}: for training data, we apply random cropping, a 224$\times$224 image resizing, and random horizontal flip for training images. For validation and test data, the 224$\times$224 image resizing is applied before center cropping images. All images are normalized using the ImageNet~\citep{deng2009imagenet} statistics. It ensures images with mean $[0.485, 0.456, 0.406]$ and standard deviation $[0.229, 0.224, 0.225]$. For datasets that do not have predefined training and test sets (i.e., datasets from Cifar10-C and DomainNet), we manually create \emph{deterministic} training and test sets following a $80\%-20\%$ split. The deterministic part is crucial to ensure no data contamination. 

\paragraph{Finetuning setup.} Our finetuning experiments follow the protocol from~\citet{dosovitskiy2021vit} with a resolution of $224 \times 224$. We optimize models with the Stochastic Gradient Descent (SGD), a momentum of 0.9, no weight decay, a cosine learning rate decay, a batch size of $512$, and gradient clipping at norm $1$. The finetuning resolution is of $224$. For each pair of dataset - configuration, we perform a sweep over $4$ learning rates, as summarized in~\cref{tab:training_details}, and conduct $3$ runs with different seeds relative to network initialization and dataloaders. For each run, we monitor the training using a validation set ($20\%$ of the training set). The final performance is the test accuracy of the checkpoint that achieves the best validation accuracy. 

\begin{table}[!h]
    \centering
       \caption{\textbf{Finetuning hyperparameters}. We report the choice of optimizer, batch size, training steps, and learning rates.}
       \scalebox{1}{
        \begin{tabular}{lccccc}
        dataset & optimizer & batch size & training steps & learning rates $\eta$\\
        \toprule[\thick pt]
        Cifar10 & SGD & 512 & $10000$ & \{$1\mathrm{e}{-3}$, $3\mathrm{e}{-3}$, $1\mathrm{e}{-2}$, $3\mathrm{e}{-2}$\} \\
        Cifar100 & SGD & 512&  $10000$& \{$1\mathrm{e}{-3}$, $3\mathrm{e}{-3}$, $1\mathrm{e}{-2}$, $3\mathrm{e}{-2}$\} \\
        Contrast & SGD & 512&  $10000$ &\{$1\mathrm{e}{-3}$, $3\mathrm{e}{-3}$, $1\mathrm{e}{-2}$, $3\mathrm{e}{-2}$\} \\
        Gaussian Noise & SGD & 512&  $10000$ &\{$1\mathrm{e}{-3}$, $3\mathrm{e}{-3}$, $1\mathrm{e}{-2}$, $3\mathrm{e}{-2}$\} \\
        Motion Blur & SGD & 512&  $10000$ &\{$1\mathrm{e}{-3}$, $3\mathrm{e}{-3}$, $1\mathrm{e}{-2}$, $3\mathrm{e}{-2}$\} \\
        Snow & SGD & 512&  $10000$ &\{$1\mathrm{e}{-3}$, $3\mathrm{e}{-3}$, $1\mathrm{e}{-2}$, $3\mathrm{e}{-2}$\} \\
        Speckle Noise & SGD & 512&  $10000$ &\{$1\mathrm{e}{-3}$, $3\mathrm{e}{-3}$, $1\mathrm{e}{-2}$, $3\mathrm{e}{-2}$\} \\
        Clipart & SGD & 512&  $20000$ & \{$3\mathrm{e}{-3}$, $1\mathrm{e}{-2}$, $3\mathrm{e}{-2}$, $6\mathrm{e}{-2}$\}\\
        Sketch& SGD & 512&  $20000$ & \{$3\mathrm{e}{-3}$, $1\mathrm{e}{-2}$, $3\mathrm{e}{-2}$, $6\mathrm{e}{-2}$\} \\
        Flowers102 & SGD & 512&  $5000$ & \{$1\mathrm{e}{-3}$, $3\mathrm{e}{-3}$, $1\mathrm{e}{-2}$, $3\mathrm{e}{-2}$\} \\
        Pets & SGD & 512&  $4000$ & \{$1\mathrm{e}{-3}$, $3\mathrm{e}{-3}$, $1\mathrm{e}{-2}$, $3\mathrm{e}{-2}$\} \\
        \end{tabular}
        }
    \label{tab:training_details}
\end{table}
\setlength{\tabcolsep}{0.3em}
\begin{table}[!h]
    \centering
    \caption{\textbf{Full finetuning results.} We report the best top-$1$ accuracy (\%) on the test set over the learning rate grid of each dataset ($\uparrow$). Each entry shows the mean and standard deviation over three finetuning runs with different seeds.}
    \scalebox{.65}{
    \begin{NiceTabular}{lccccccccccc}
     Dataset & \rotatebox{45}{Cifar10} & \rotatebox{45}{Cifar100}& \rotatebox{45}{Contrast}& \rotatebox{45}{Gaussian Noise}& \rotatebox{45}{Motion Blur}& \rotatebox{45}{Snow}& \rotatebox{45}{Speckle Noise}& \rotatebox{45}{Clipart}& \rotatebox{45}{Sketch}& \rotatebox{45}{Flowers102}& \rotatebox{45}{Pets} \\
     \toprule[\thick pt] 
    \textit{full finetuning} & $\text{99.02}_{\tiny \pm \text{0.02}}$ & $\text{92.74}_{\tiny \pm \text{0.05}}$ & $\text{97.23}_{\tiny \pm \text{0.18}}$ & $\text{87.14}_{\tiny \pm \text{1.16}}$ & $\text{94.67}_{\tiny \pm \text{0.14}}$ & $\text{95.42}_{\tiny \pm \text{0.13}}$ & $\text{89.58}_{\tiny \pm \text{0.43}}$ & $\text{78.50}_{\tiny \pm \text{0.49}}$ & $\text{71.30}_{\tiny \pm \text{0.26}}$ & $\text{99.15}_{\tiny \pm \text{0.05}}$ & $\text{94.57}_{\tiny \pm \text{0.29}}$ \\
    \end{NiceTabular}
    }
    \label{tab:finetuning_results}
\end{table}

\section{Additional experiments}
\begin{figure}[!h]
    \centering
    \includegraphics[width=\linewidth]{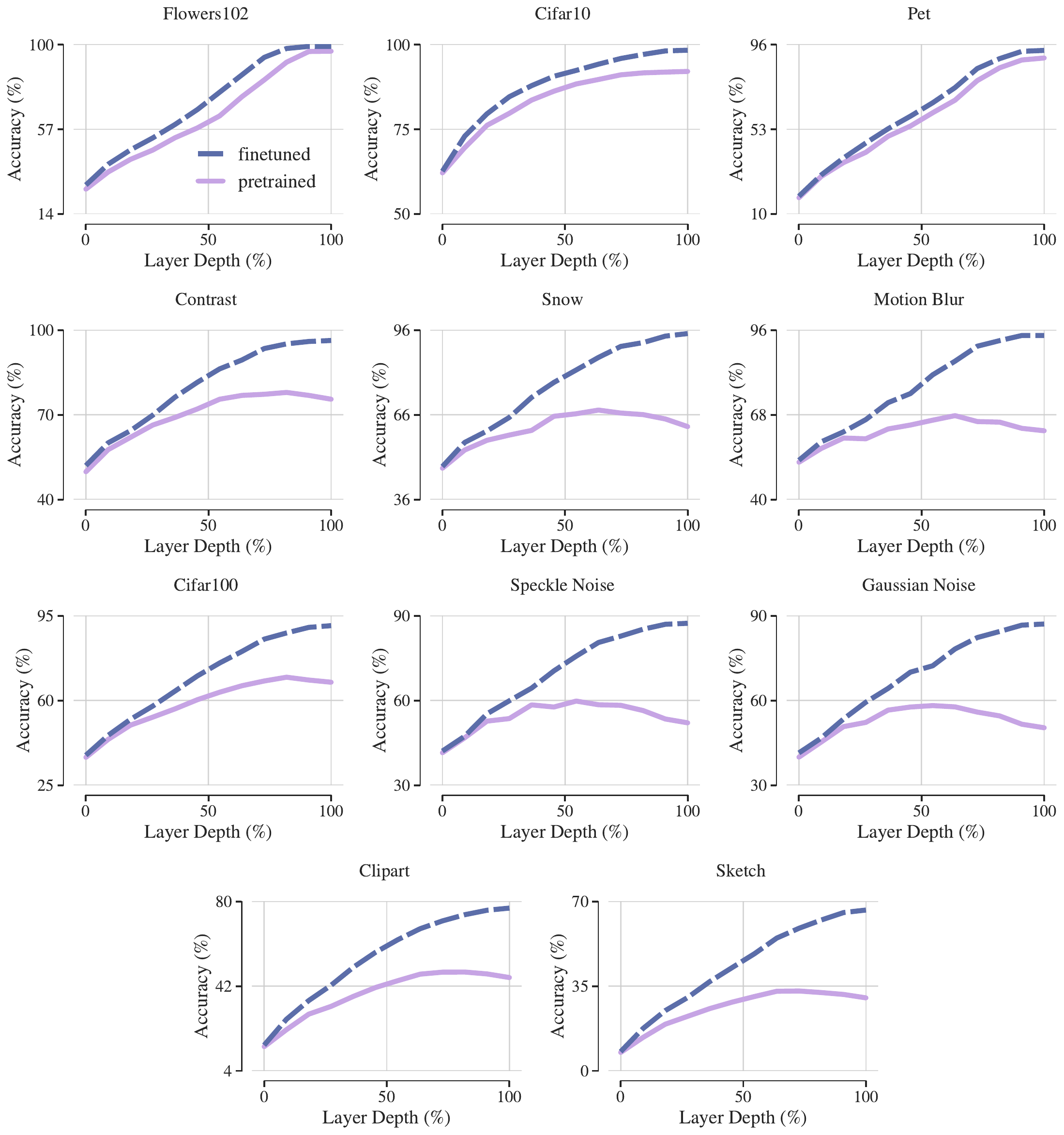}
    \caption{\textbf{Layer by layer.} Evolution of the linear probing performance across the layers of an $86$M ViT pretrained on ImageNet. The solid line denotes the model only pretrained, and the dashed line denotes the model finetuned on the dataset at hand. From \textbf{left} to \textbf{right}, the shift between the pretraining and the downstream data increases. The stronger the shift, the worse the final layers.}
    \label{fig:layer_by_layer_all}
\end{figure}

\begin{figure}[!h]
    \centering
    \includegraphics[width=\linewidth]{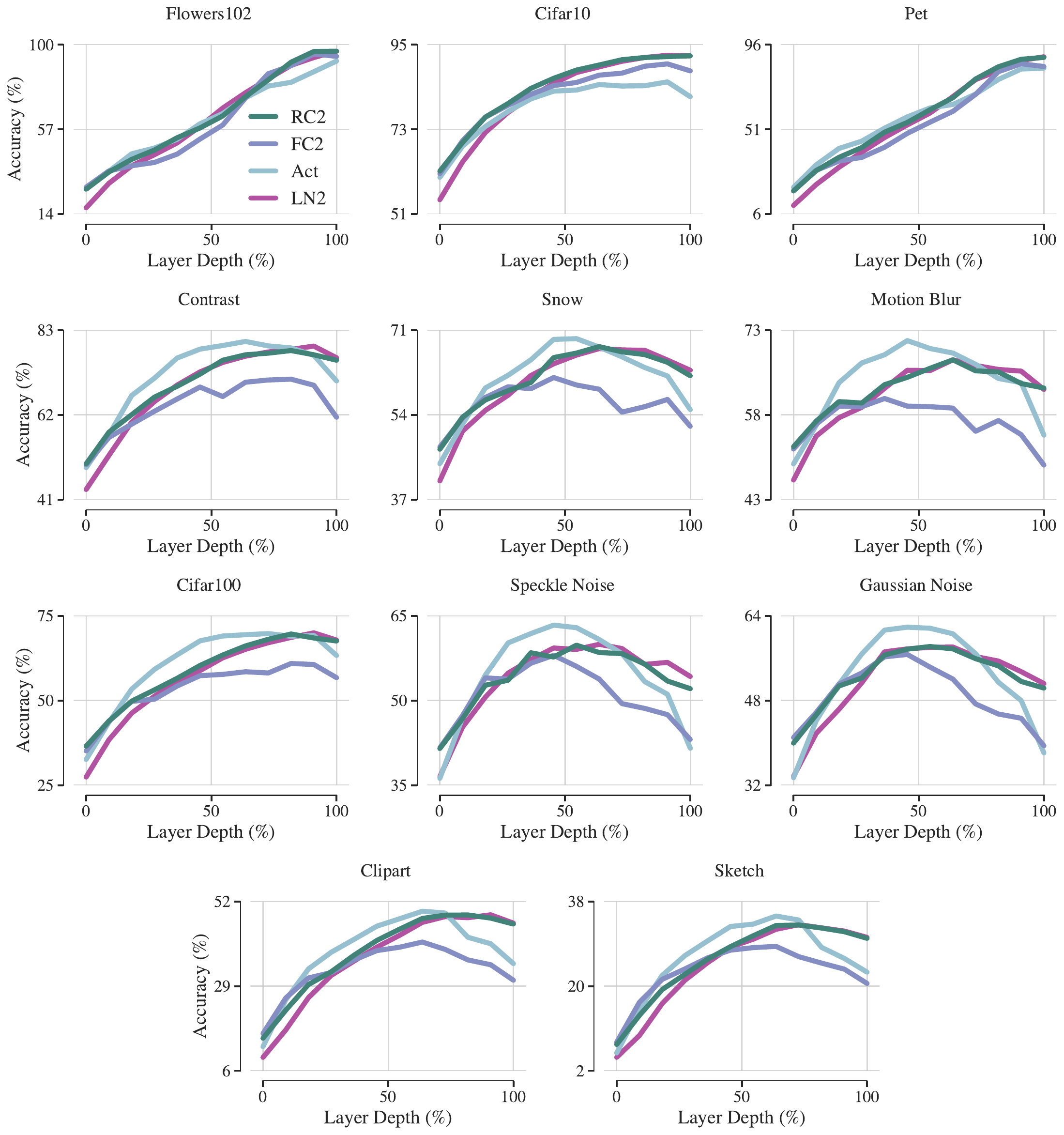}
    \caption{\textbf{Layer by layer, module by module.} Evolution of the linear probing performance of transformer modules across the layers of an $86$M ViT pretrained on ImageNet. From \textbf{left} to \textbf{right}, the shift between the pretraining and the downstream data increases.}
    \label{fig:module_by_module_all}
\end{figure}

We display in~\cref{fig:layer_by_layer_all,fig:module_by_module_all} the additional results on all benchmarks related to the experiments of~\cref{sec:layer_by_layer,sec:module_by_module}, respectively.

\end{document}